\documentclass{article}
\usepackage{ijcai26}

\usepackage{times}
\usepackage{soul}
\usepackage{url}
\usepackage[hidelinks]{hyperref}
\usepackage[utf8]{inputenc}
\usepackage[small]{caption}
\usepackage{graphicx}
\usepackage{amsmath}
\usepackage{amsthm}
\usepackage{booktabs}
\usepackage{algorithm}
\usepackage[noend]{algorithmic}
\usepackage[switch]{lineno}
\usepackage{natbib}

\newtheorem{example}{Example}

\title{A Survey on the Verification of Reinforcement Learning Policies}

\author{
Luca Marzari$^1$,
Ezio Bartocci$^1$
\And
Enrico Marchesini$^2$\\
\affiliations
$^1$TU Wien, Vienna, Austria\\
$^2$Massachusetts Institute of Technology, Cambridge (MA), USA\\
\emails
luca.marzari@tuwien.ac.at,
ezio.bartocci@tuwien.ac.at,
emarche@mit.edu
}

\usepackage{amsmath}
\usepackage{amssymb}
\usepackage{mathrsfs}
\usepackage{amscd}
\usepackage{amsfonts}

\usepackage{lipsum}  

\usepackage{enumitem}   % reduce itemize space
\usepackage{wrapfig}    % wrap figures
\usepackage{xcolor}
\usepackage{etoolbox}
\usepackage{tcolorbox}
\usepackage{makecell}
\usepackage{pifont}
\usepackage{graphics,color}
\usepackage{dsfont}
\usepackage[table]{xcolor}
\usepackage[bottom]{footmisc} 
\usepackage{multirow}
\usepackage{threeparttable}
\usepackage{float}

\newcommand{\cmark}{\textcolor{green!70!black}{\checkmark}}
\newcommand{\xmark}{\textcolor{red}{\times}}

\newtheorem{definition}{\textbf{Definition}}

\usepackage{tikz,ifthen}
\usetikzlibrary{arrows,trees,backgrounds,automata,shapes,decorations,plotmarks,fit,calc,positioning,shadows,chains}

\tikzset{
  weight/.style={
    font=\scriptsize,
    fill=white,
    inner sep=1pt,
    text opacity=1
  }
}

\definecolor{color0}{RGB}{240, 78, 64} %output
\definecolor{color4}{RGB}{60, 220, 125} %input
\definecolor{color6}{RGB}{120, 100, 200} %hidden
\definecolor{color7}{RGB}{107, 100, 200} %hidden

\definecolor{nnedgecolor}{RGB}{90,90,90}
\tikzstyle{every pin edge}=[<-,shorten <=1pt]
\tikzstyle{every path}=[draw=color7!50]
\tikzstyle{neuron}=[circle,fill=black!25,minimum size=17pt,inner sep=0pt]
\tikzstyle{input neuron}=[neuron, fill=color4]
\tikzstyle{output neuron}=[neuron, fill=color0]
\tikzstyle{hidden neuron}=[neuron, fill=color6]
\tikzstyle{annot} = [text width=4em, text centered]
\tikzstyle{nnedge} = [-{stealth},shorten >=0.1cm, shorten <=0.05cm,line 
width=0.8pt,nnedgecolor]
\tikzstyle{nnedge_t} = [-{dashed},shorten >=0.1cm, shorten <=0.05cm,line 
width=0.8pt,nnedgecolor]
\usetikzlibrary{calc}

\begin{document}

\maketitle

\begin{abstract}
Reinforcement learning (RL) is increasingly applied in complex, safety-critical domains, yet the lack of rigorous behavioral guarantees for neural network-based policies remains a major barrier to deployment. Recent advances in policy expressiveness and scale have intensified this challenge, leading to a rapidly growing but conceptually fragmented body of work on RL policy verification.
This survey provides a unifying perspective on RL verification methods. We introduce a taxonomy that clarifies relationships among existing approaches along three axes: verification paradigm (formal versus probabilistic), temporal scope (step-wise versus multi-step), and guarantees strength. Beyond taxonomy, we unify underlying theoretical foundations, make implicit assumptions and limitations explicit, and identify emerging directions.

% such as multi-agent verification and verification methods for modern policy architectures.

% Reinforcement Learning (RL) has successfully conquered complex domains, ranging from Go to challenging robotic medical control. However, its deployment in safety-critical systems is stalled by the "black box" nature of neural policies. This survey bridges the gap between verification methods and reinforcement learning. Specifically, we present a unified taxonomy for post-training verification of RL Agents. We categorize methods across three axes: (1) verification paradigm (formal deterministic or probabilistic), (2) RL verification (robustness or safety), and (3) temporal horizon (step-wise robustness or safety verification vs. infinite-horizon reachability). We highlight emerging trends in verification methods for risk quantification and the verification of modern architectures, such as Transformers.
    
\end{abstract}

\section{Introduction}
\label{sec:introduction}

Over the past decade, reinforcement learning (RL) has emerged as a powerful framework for sequential decision making in complex, high-dimensional environments \citep{DRL}. Recent advances in expressiveness and scalable training have accelerated interest in deploying RL agents in safety-critical domains such as autonomous driving and energy systems \citep{vinitsky_drive, marchesini2025rl2grid, marl2grid_iclr2026}. Despite these advances, the deployment of RL systems in high-stakes settings remains limited by the lack of rigorous \emph{behavioral} guarantees. RL policies are typically realized as deep neural networks (DNNs) with complex, non-linear decision boundaries, which can exhibit unsafe behavior under rare conditions, distribution shift, or interaction effects \citep{adversarial, aydeniz_2025}. As empirical validation alone cannot rule out such failures \citep{aamas23, TIST}, formal and systematic verification becomes essential for safety-critical RL applications.

This requirement has given rise to the application of \emph{formal verification of deep neural networks} methods to RL, which seek to provably certify post-training, that a policy satisfies desired behavioral properties prior to deployment \citep{LiuSurvey}. Unlike supervised learning, RL induces correctness properties that fundamentally depend on interaction, feedback, and distribution shift under policy execution. An RL policy does not operate on a fixed input distribution, but actively shapes the state distribution it encounters through closed-loop control, leading to error accumulation, rare-event failures, and safety violations that may only arise after long horizons (e.g., a policy that is locally robust at every time step may still violate safety through compounding errors over long horizons) \citep{Sutton1998}. As a result, verification objectives in RL cannot be reduced to input-output robustness alone, but must account for temporal dependence, policy-environment coupling, and deployment-time uncertainty. These characteristics fundamentally shape both what can be verified and which abstractions are meaningful. In detail, we verify properties including safety guarantees (e.g., collision avoidance), robustness guarantees (e.g., resilience to noise, or adversarial perturbations), and reachability guarantees (e.g., task completion). Providing these certificates has become increasingly challenging as RL policies grow in expressiveness and scale. 

\begin{table*}[!ht]
\centering
\begin{threeparttable}
\resizebox{\textwidth}{!}{%
\begin{tabular}{@{}l l l c l c c c c c@{}}
\toprule
\textbf{Paradigm} & \textbf{Tool / Paper} & \textbf{Approach / Solver} & \textbf{Horizon} & \textbf{Arch.} & 
\textbf{Rob.} & \textbf{Safe.} & \textbf{Enum.} & 
\textbf{Guarantee} & \textbf{RL-Ready} \\ 
\midrule

% ================= FORMAL =================
\multirow{10}{*}{\textbf{Formal}} 

& \multirow{2}{*}{$\alpha,\beta$-CROWN \citep{autolirpa}}
& \multirow{2}{*}{Bound Prop. / Linearization} 
& Step-wise & MLP, CNN, RNN & $\boldsymbol{\cmark}$ & $\boldsymbol{\cmark}$ & $\boldsymbol{\xmark}$ 
& Sound \& Complete\tnote{*} & Yes \\

& & & Step-wise & Transformer & $\boldsymbol{\cmark}$ & $\boldsymbol{\xmark}$ & $\boldsymbol{\xmark}$ 
& Sound & No \\ \cmidrule(l){2-10}

& nnenum \citep{nnenum}, PyRAT \citep{pyrat}
& \multirow{2}{*}{Abs. Int. / Reachability} 
& Step-wise & MLP, CNN & $\boldsymbol{\cmark}$ & $\boldsymbol{\xmark}$ & $\boldsymbol{\xmark}$ 
& Sound  & Yes \\

& NNV 2.0 \citep{NNV2.0} & 
& Step-wise & MLP, Neural-ODE & $\boldsymbol{\cmark}$ & $\boldsymbol{\cmark}$ & $\boldsymbol{\xmark}$ 
& Sound  & Yes \\ \cmidrule(l){2-10}

& ModelVerification.jl \citep{ModelVerification} & Bound Prop. / Linearization
& Step-wise & MLP, CNN, Neural-ODE & $\boldsymbol{\cmark}$ & $\boldsymbol{\cmark}$ & $\boldsymbol{\xmark}$ 
& Sound \& Complete\tnote{*}  & Yes \\ \cmidrule(l){2-10}

& Marabou (Reluplex extension) \citep{Marabou2.0}
& \multirow{2}{*}{SMT / MIP Solving} 
& Step-wise & MLP, CNN & $\boldsymbol{\cmark}$ & $\boldsymbol{\cmark}$ & $\boldsymbol{\xmark}$ 
& Sound \& Complete  & Yes \\

& NeuralSAT \citep{NeuralSAT}
&  & Step-wise & MLP & $\boldsymbol{\cmark}$ & $\boldsymbol{\cmark}$ & $\boldsymbol{\xmark}$ 
& Sound \& Complete  & Yes \\ \cmidrule(l){2-10}

& U/L-Dist \citep{kofnovexact}
& Bound Prop. / Reachability & Step-wise & MLP,CNN & $\boldsymbol{\cmark}$ & $\boldsymbol{\xmark}$ & $\boldsymbol{\xmark}$ 
& Sound \& Complete\tnote{*} & Partial \\ \cmidrule(l){2-10}

& Neural Lyapunov \citep{yang2024lyapunov}
& Control Theory 
& Multi-Step & Neural CBF & $\boldsymbol{\xmark}$ & $\boldsymbol{\cmark}$ & $\boldsymbol{\xmark}$ 
& Sound \& Complete\tnote{*}  & Yes \\ \cmidrule(l){2-10}

& INVPROP \citep{INVPROP}
& Bound Prop. / Linearization 
& Step-wise & MLP & $\boldsymbol{\xmark}$ & $\boldsymbol{\cmark}$ & $\boldsymbol{\cmark}$ 
& Sound  & Yes \\ \cmidrule(l){2-10}

& Exact Enum \citep{matoba2020exact}
& Preimage Enumeration 
& Step-wise & MLP & $\boldsymbol{\xmark}$ & $\boldsymbol{\cmark}$ & $\boldsymbol{\cmark}$ 
& Sound \& Complete & Yes \\

\midrule
% ================= PROBABILISTIC =================
\multirow{9}{*}{\textbf{Probabilistic}} 

& PT-LiRPA \citep{ptlirpa}
& Probabilistic Linear Relaxation 
& Step-wise & MLP, CNN & $\boldsymbol{\cmark}$ & $\boldsymbol{\cmark}$ & $\boldsymbol{\xmark}$ 
& Probabilistic  & Yes \\ \cmidrule(l){2-10}

& Online CBF \citep{ral}
& Runtime Monitoring 
& Multi-Step & MLP + CBF & $\boldsymbol{\xmark}$ & $\boldsymbol{\cmark}$ & $\boldsymbol{\cmark}$ 
& Probabilistic  & Yes \\ \cmidrule(l){2-10}

& PL-MDP \citep{Abate}
& Temporal Logic 
& Multi-step & MLP (abstracted) & $\boldsymbol{\xmark}$ & $\boldsymbol{\cmark}$ & $\boldsymbol{\xmark}$ 
& Probabilistic  & Partial \\ \cmidrule(l){2-10}

% & Safety Index \citep{SafetyIndexGP}
% & Synthesis 
% & Step-wise & MLP + GP & $\boldsymbol{\xmark}$ & $\boldsymbol{\cmark}$ & $\boldsymbol{\xmark}$ 
% & Probabilistic & Medium & Yes \\ \cmidrule(l){2-11}

& FSC-based \citep{carr2021verifiable}
& Synthesis 
& Step-wise & MLP, RNN & $\boldsymbol{\xmark}$ & $\boldsymbol{\cmark}$ & $\boldsymbol{\xmark}$ 
& Probabilistic  & Partial \\ \cmidrule(l){2-10}

& PREMAP \citep{zhang2025premap}
& Bound Prop. / Linearization 
& Step-wise & MLP & $\boldsymbol{\xmark}$ & $\boldsymbol{\cmark}$ & $\boldsymbol{\cmark}$ 
& Sound / Probabilistic  & Partial \\ \cmidrule(l){2-10}

& Prob. Enum \citep{RFprove}
& Abs. Int. / Reach. + Sampling 
& Step-wise & MLP & $\boldsymbol{\xmark}$ & $\boldsymbol{\cmark}$ & $\boldsymbol{\cmark}$ 
& Probabilistic  & Yes \\

\bottomrule
\end{tabular}%
}
\begin{tablenotes}
\tiny
\item[*] Not for all the methods, completeness holds only when combined with exhaustive branching or linear programming solvers.
\end{tablenotes}
\end{threeparttable}
\vspace{-5pt}
\caption{A unified taxonomy of verification methods for deep RL, highlighting trade-offs between guarantee strength and expressivity across step-wise, multi-step, probabilistic, and enumeration-based formulations.}
\label{tab:taxonomy_updated}
\vspace{-5pt}
\end{table*}

\begin{figure}[t!]
    \centering
    \includegraphics[width=.9\linewidth]{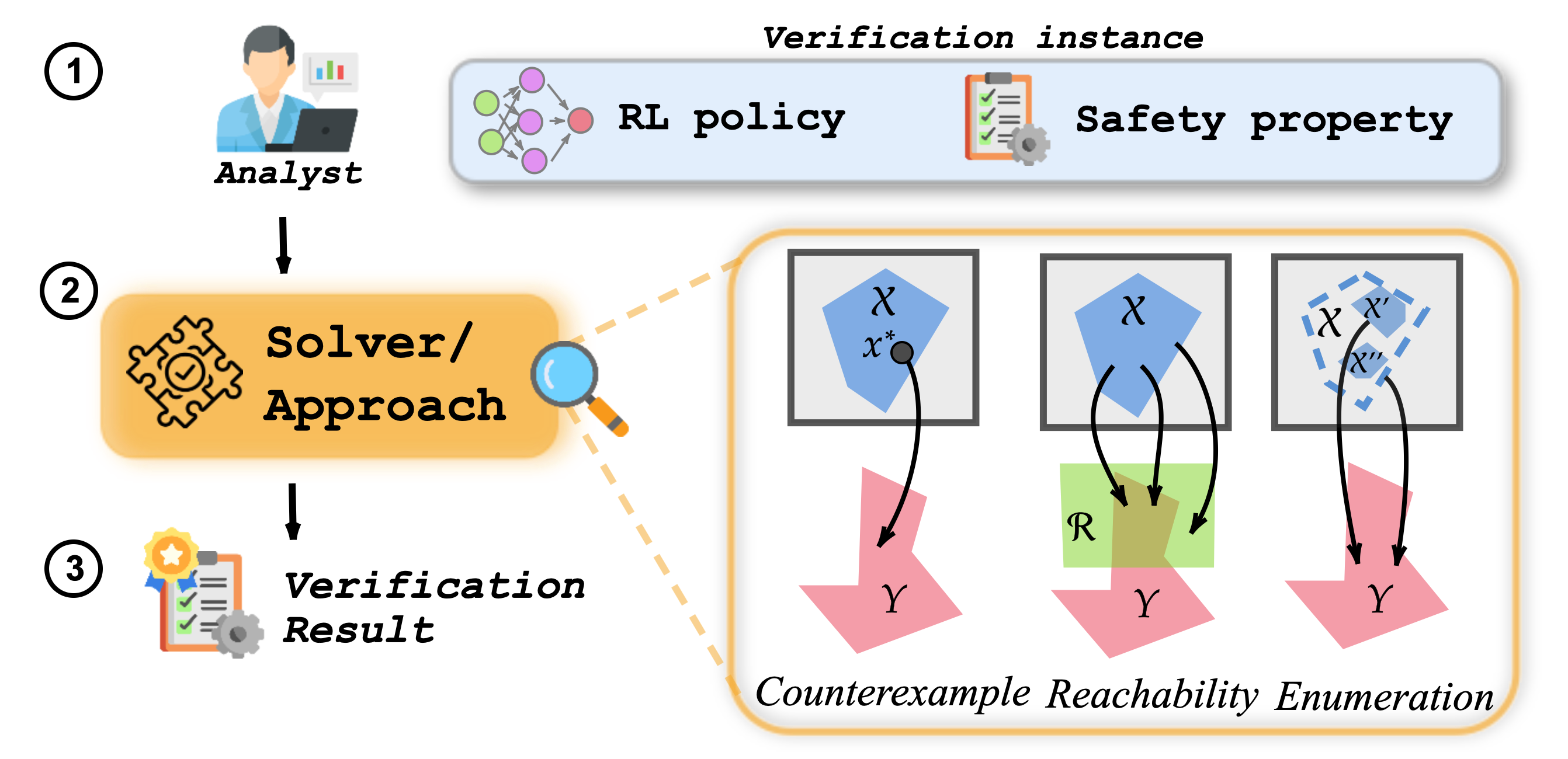}
    \vspace{-5pt}
    \caption{Pipeline for post-training verification of RL policies, highlighting solver-based approaches used to establish safety guarantees.}
    \label{fig:pipeline}
    \vspace{-6pt}
\end{figure}

Fig. \ref{fig:pipeline} summarizes the post-training RL verification pipeline, where a safety property and trained policy are analyzed by a verification solver to either produce a counterexample, establish a reachability-based guarantee, or enumerate unsafe regions of the state space. Drawing on formal methods, control theory, and statistical analysis, a wide range of approaches exists to certify policy behavior after training. These include deterministic techniques that provide sound guarantees via satisfiability (SAT) \citep{SAT} and reachability analysis \citep{lomuscio2017approach}, probabilistic methods \citep{valiant1984theory} that quantify risk in stochastic environments, and hybrid neuro-symbolic approaches that combine learned policies with control theory \citep{zhao2020synthesizing}. Despite the growing volume of results, the literature remains conceptually fragmented. Methods are often developed under different assumptions, target different notions of correctness, and operate over different temporal scopes, making it difficult to compare approaches or understand their fundamental relationships.\\
\textbf{Contributions.}
We address this fragmentation by introducing a unified taxonomy of post-training verification methods for neural network-based policies. We organize existing approaches along three core axes: the verification paradigm, the temporal scope of properties, and the guarantees strength of different solvers. Beyond taxonomy, we formalize the main RL verification problems using a unified notation and provide clear, illustrative examples to clarify the underlying ideas. We also analyze the limitations of existing methods and the key trade-offs among expressiveness and soundness.
%, and scalability quantified using three levels—\emph{low}, \emph{medium}, and \emph{high}, based on the size of the neural network models and the dimensionality of the input or state space, as reflected in standard verification benchmarks. 
%Specifically, \emph{Low scalability} methods are typically limited to networks with at most $10^4$--$10^5$ parameters and low-dimensional inputs (up to $O(10)$ dimensions). \emph{Medium scalability} methods can handle networks with up to $10^6$--$10^7$ parameters and input dimensions in the tens to low hundreds.  Finally, \emph{High scalability} methods scale to networks with tens to hundreds of millions of parameters and effective input dimensions in the thousands, as demonstrated by benchmarks involving large convolutional or vision-based architectures. 
Finally, we discuss open challenges in verifying modern policy architectures, multi-agent systems, and outline directions for extending verification techniques to these settings.

\section{Preliminaries}
\label{sec:rl_background}
%We start by briefly reviewing the RL formalism and execution semantics that underlie policy verification.
Reinforcement learning is commonly formalized as a Markov decision process (MDP) $\langle \mathcal{S}, \mathcal{A}, P, R, \gamma \rangle$, where $\mathcal{S}$ and $\mathcal{A}$ denote finite state and action spaces, respectively, $P : \mathcal{S} \times \mathcal{A} \times \mathcal{S} \to [0,1]$ specifies the state transition dynamics, $R : \mathcal{S} \times \mathcal{A} \to \mathbb{R}$ is the reward function, and $\gamma \in [0,1)$ is the discount factor. An agent’s behavior is determined by a policy mapping states to actions, which is parameterized by a DNN for the context of this survey. When deployed, the policy induces a distribution over state-action trajectories. 

%From a verification perspective, behavioral guarantees for RL are thus properties of sequential behaviors rather than isolated inputs. %: safety or robustness violations may arise only through the cumulative effect of multiple decisions. 
Most verification methods studied here reason over abstractions of this policy-environment loop, differing in how explicitly they model feedback, uncertainty, and temporal dependence. Understanding which aspects of RL execution are preserved or abstracted away is thus essential for interpreting the guarantees provided by different verification methods.

\section{A Unified Taxonomy on RL Verification}
\label{sec:taxonomy}

While many underlying solver techniques originate from classical neural network verification, their application to RL raises distinct semantic challenges due to feedback, temporal dependence, and interaction with uncertain environments. Table \ref{tab:taxonomy_updated} organizes RL verification approaches along three main axes, plus an ``RL readiness" one. \emph{Paradigm} characterizes whether the method is formal or probabilistic.
\emph{Temporal scope} distinguishes step-wise methods, which analyze a single policy evaluation under bounded inputs, from multi-step methods, which reason about trajectory-level or closed-loop behavior. \emph{Guarantees} indicates the strength of the assurance provided. 
%\emph{Scalability and RL readiness} then capture practical applicability: scalability reflects how methods scale with network size and input dimensionality, while RL readiness indicates whether a method can be applied directly to trained RL policies without substantial abstraction or reformulation.
\emph{RL readiness} then capture practical applicability, indicating whether a method can be applied directly to trained RL policies without substantial abstraction or reformulation.

\paragraph{Reader’s Roadmap.}
This section progresses through increasingly expressive RL verification abstractions. We begin with step-wise analysis (\ref{sec:step-wise}), reasoning about 1-step individual policy decisions. We then consider multi-step formulations (\ref{sec:multi-step}), capturing closed-loop dynamics and temporal properties at higher computational cost. Hence, enumeration-based approaches (\ref{sec:enumeration}) that characterize unsafe regions of the state space. Table \ref{tab:taxonomy_updated} provides a unifying reference.

\subsection{Step-wise RL Verification}
\label{sec:step-wise}
Step-wise RL verification \emph{asks whether a policy selects an unsafe action at a given state}, abstracting away future dynamics. Yet, it underpins most scalable approaches and remains the most mature class of RL verification tools.

% Step-wise verification is thus characterized by a \emph{local temporal scope}, applies to both \emph{formal} and \emph{probabilistic} verification paradigms, and primarily targets \emph{robustness} and \emph{safety} objectives at the level of individual decisions.

\textit{Problem Formulation.}
Consider a trained RL policy represented by a DNN $f: \mathcal{I} \rightarrow \mathcal{O}$, where inputs encode the agent’s state and outputs represent action values, probabilities, or control commands. 
The step-wise verification problem is specified by a tuple $\mathcal{T} = \langle f, \mathcal{X}, \mathcal{Y} \rangle$, where the precondition $\mathcal{X} \subseteq \mathcal{I}$ defines a set of admissible inputs (typically a neighborhood around a nominal state capturing uncertainty due to sensor noise, or state estimation error). The postcondition $\mathcal{Y} \subseteq \mathcal{O}$ encodes unsafe actions or violations of control constraints. The pair $(\mathcal{X}, \mathcal{Y})$ thus defines a \emph{local behavioral property} to verify for the policy, evaluated independently of the environment \citep{LiuSurvey}.

\begin{figure}[t]
        \centering
        \includegraphics[width=.85\linewidth]
        {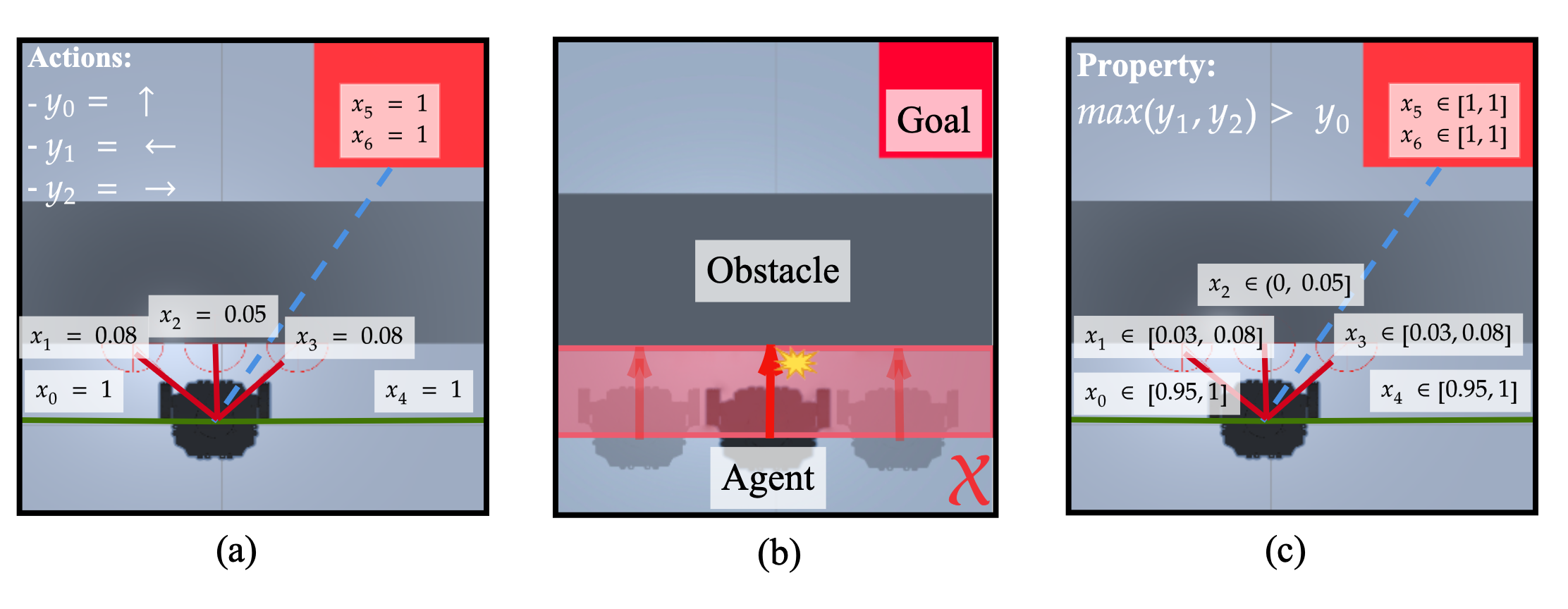}
        \vspace{-6pt}
        \caption{Explanatory behavioral property. (a) Unsafe situation to be verified. (b) Encoding of the unsafe region using $\mathcal{X}$ as geometric set. (c) Property encoded using intervals and predicate on the output.}
        \vspace{-5pt}
        \label{fig:safety_prop}
    \end{figure}

\begin{example}[Local behavioral property]
Consider the navigation task in Fig.~\ref{fig:safety_prop}, where a policy maps lidar and goal features to actions $\{y_0,y_1,y_2\}$ (forward, left, right). When an obstacle is directly ahead, safety requires that the forward action is never selected under bounded sensor noise. This is encoded by defining $\mathcal{X}$ as an $\ell_\infty$ ball around the nominal obstacle state and the postcondition $\mathcal{Y}$ as $\max(y_1,y_2) > y_0$. The property is violated if any state in $\mathcal{X}$ induces $y_0$.
\end{example}

Notably, in step-wise RL verification, robustness and safety may coincide: $\mathcal{X}$ defines a local perturbation region around a given state, and verifying that no input in $\mathcal{X}$ induces an unsafe action may be equivalent to certifying adversarial robustness of the policy at that state. This equivalence enables a standardized encoding using a specification language such as VNN-LIB \citep{VNNlib}, where $\mathcal{X}$ is specified by linear input constraints and $\mathcal{Y}$ by linear output predicates (e.g., action-ranking constraints). Step-wise RL verification can thus be addressed directly by off-the-shelf neural network verifiers without modeling the environment, allowing methods ranging from exact mixed-integer programming \citep{MILP} to scalable over-approximate techniques such as abstract interpretation \citep{Ai2} and linear relaxation \citep{fastlin, crown} to be applied in a unified and reproducible manner.

This equivalence admits two definitions for step-wise verification: an explicit search for a violating input in $\mathcal{X}$ with SAT analysis \citep{Reluplex, Marabou2.0, MIP}, or the computation of conservative output bounds over all inputs in $\mathcal{X}$ with the reachability-based formulation \citep{lomuscio2017approach,reluval,fastlin, kofnovexact}, which we discuss next.

\paragraph{Satisfiability Setting.}
In SAT-based verification (Def. \ref{def:sat_rl_verification}), the policy and the negation of the safety property are encoded as a set of logical constraints \citep{NeuralSAT}. Verification reduces to checking whether there exists an input in $\mathcal{X}$ such that the corresponding output lies in $\mathcal{Y}$. When a solution exists, the verifier returns a counterexample, corresponding to the state where the policy violates the behavioral property.

\begin{tcolorbox}
  \vspace{-0.1cm}

\begin{definition}[\textsc{SAT-Based RL Verification}]
\label{def:sat_rl_verification}
\phantom{a}

{\bf Input}: A tuple $\mathcal{T} = \langle f, \mathcal{X}, \mathcal{Y} \rangle$

{\bf Output}: $\texttt{violate} \iff \exists x \in \mathcal{X} \;\vert\; f(x) \in \mathcal{Y}$
\end{definition}
\end{tcolorbox}

SAT-based methods are particularly valuable for falsification and debugging of learned policies. However, their worst-case computational complexity grows exponentially with network size (non-linearities), limiting scalability. In particular, this is an NP-hard problem \citep{Reluplex}, which motivates the development of over-approximation techniques and branch-and-bound strategies \citep{bab}.

\paragraph{Reachability Setting.} This alternative strategy (Def. \ref{def:reachability_rl_verification}) propagates bounds through the policy network to compute an over-approximation of the reachable output set $\mathcal{R}(\mathcal{X})$ (e.g., by using interval bound propagation \citep{lomuscio2017approach}). When $\mathcal{R}(\mathcal{X})$ is fully contained within the unsafe output region $\mathcal{Y}$, the policy violates the behavioral property.

\begin{tcolorbox}
  \vspace{-0.1cm}

\begin{definition}[\textsc{Reachability-Based RL Verification}]
\label{def:reachability_rl_verification}
\phantom{a}

    {\bf Input}: A tuple $\mathcal{T} = \langle f, \mathcal{X}, \mathcal{Y} \rangle$
    
    {\bf Output}: $\texttt{violate} \iff \mathcal{R}(\mathcal{X}) \subseteq \mathcal{Y}$
\end{definition}
\end{tcolorbox}

\begin{figure}[b!]
    \centering
    \vspace{-5pt}
    \includegraphics[width=0.65\linewidth]{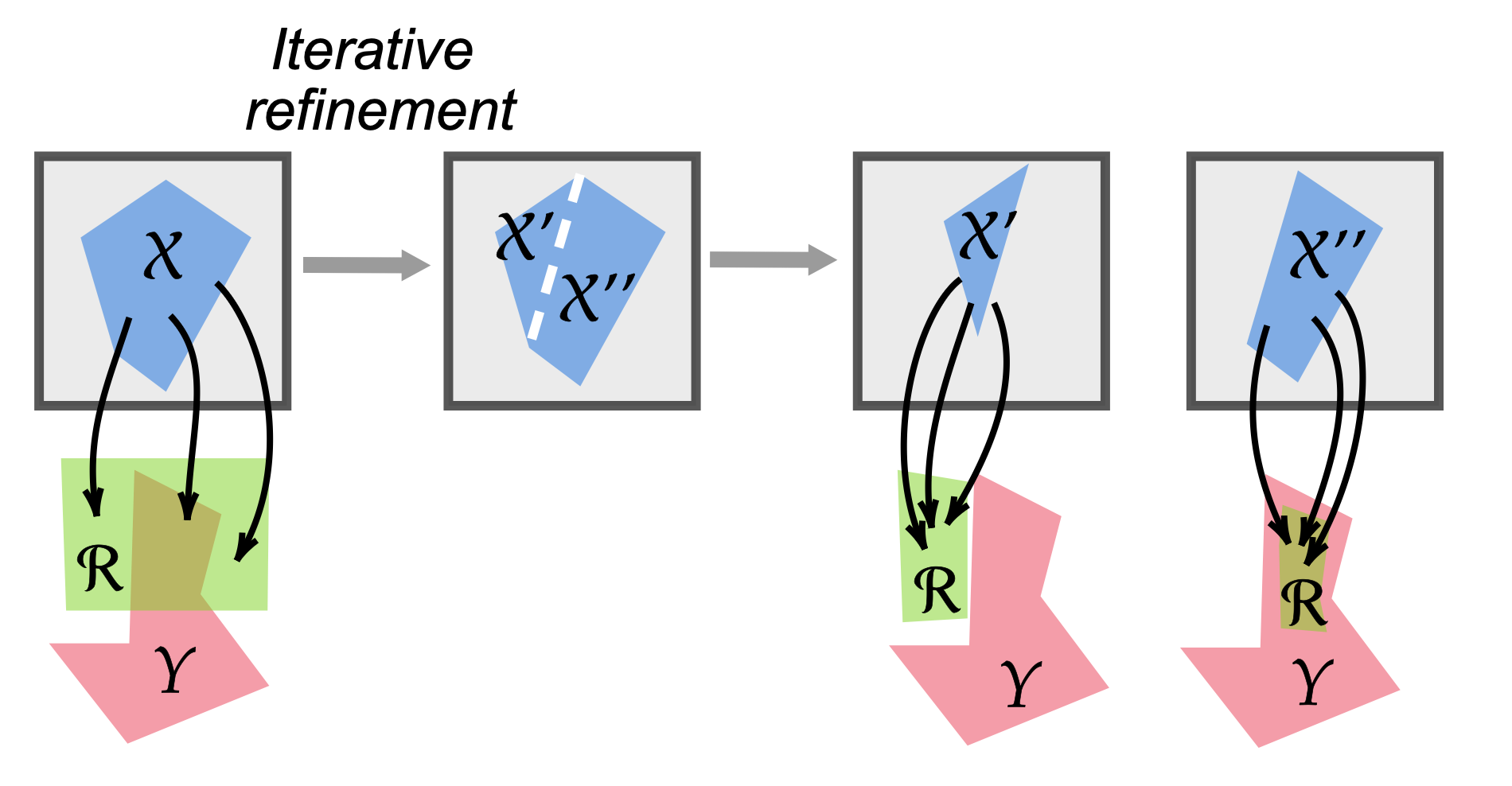}
    \vspace{-5pt}
    \caption{BaB refinement. (Adapted from \citep{eProve}.)}
    \label{fig:bab}
    \vspace{-5pt}
\end{figure}

From an RL perspective, reachability-based verification is attractive because it provides guarantees that hold uniformly over regions of the state space, rather than individual counterexamples \citep{nnenum,pyrat,NNV2.0}. This is particularly well suited to RL settings where uncertainty in $\mathcal{X}$ reflects epistemic sources (e.g., perception noise, or state estimation error) rather than adversarial perturbations. In such cases, region-level guarantees better align with how policies are deployed and evaluated. The main challenge is over-approximation: conservative reachable-set bounds may be too coarse to certify or refute a property. A common remedy is domain refinement via branch-and-bound, where the input region is recursively split using heuristic criteria \citep{reluval}, yielding tighter output bounds on smaller subregions until a conclusive result is obtained (Fig.~\ref{fig:bab}). However, in the worst case, this process requires an exponential number of verification calls, reflecting the NP-hardness of verifying piecewise-linear neural networks.

Modern verifiers also combine linear relaxations (e.g., linear relaxation-based perturbation analysis (LiRPA) \citep{autolirpa}) with refined branching strategies that split either the input space or internal activation patterns (e.g., $\alpha,\beta$-CROWN \citep{crown,acrown,bcrown}, ModelVerification.jl \citep{ModelVerification}). In RL, these refinements tighten bounds on policy outputs over local neighborhoods of states, improving the precision of action-level guarantees. However, the resulting scalability-precision trade-off is especially acute for high-dimensional RL policies, motivating a probabilistic perspective.

\paragraph{Probabilistic Step-Wise Verification.} 
To improve scalability, probabilistic methods estimate the likelihood that an RL policy selects an unsafe action under uncertainty, rather than requiring safety for all inputs in $\mathcal{X}$. Early approaches, such as PROVEN \citep{proven} and randomized smoothing \citep{randomizedSmoothing}, bound violation probabilities under input perturbation distributions but rely on restrictive noise assumptions or focus on classification. More recently, \texttt{PT-LiRPA} \citep{ptlirpa} extends linear relaxation with probabilistic reasoning to bound the probability of unsafe actions within an uncertainty region without assuming a specific noise model, enabling scalable verification of RL policies.
Despite their maturity and strong tooling support, step-wise methods cannot reason about closed-loop execution or long-horizon effects, motivating the multi-step and enumeration-based paradigms discussed next.

\subsection{Multi-step RL Verification}
\label{sec:multi-step}
Multi-step RL verification \emph{asks whether a policy can incur unsafe behavior over time}. Such properties are inherently temporal and depend on closed-loop trajectories, but their expressiveness comes at a high computational cost: modeling dynamics, temporal specifications, and long-horizon uncertainty significantly limits scalability. As a result, existing approaches rely on abstractions and relaxations, giving rise to several distinct multi-step verification paradigms.

\textit{Problem Formulation.} Multi-step verification generalizes step-wise analysis by reasoning about the interaction between the policy and the environment over finite or infinite horizons.
The objective is to determine whether any closed-loop trajectory can reach an unsafe state, shifting the object of analysis from the policy in isolation to the policy-environment system. A trajectory $\tau = \{s_0,a_0,s_1,a_1,\ldots,s_t\}$ is simply a sequence of state-action interactions. When the transition dynamics are known or can be conservatively approximated, multi-step RL verification is a trajectory-level reachability problem that verifies a \emph{trajectory-level behavioral property}.

\begin{example}[Trajectory-level behavioral property]
Consider the navigation task in Fig.~\ref{fig:safety_prop}, where an agent follows a fixed policy in a known environment with deterministic dynamics. Let $\mathcal{S}_{\mathrm{unsafe}}\subseteq\mathcal{S}$ denote collision states. Given an initial set $\mathcal{X}_0\subseteq\mathcal{S}$, safety requires that no closed-loop trajectory reaches $\mathcal{S}_{\mathrm{unsafe}}$, i.e.,
$\forall s_0\in\mathcal{X}_0,\ \forall t\ge 0,\ s_t\notin\mathcal{S}_{\mathrm{unsafe}}$. The property is violated if some $s_0\in\mathcal{X}_0$ reaches an unsafe state.
\end{example}

A representative approach is the reachability analysis of neural feedback loops proposed by \cite{everett2021reachability}, which embeds the policy within a dynamical system and propagates over-approximations of reachable states over time using linearization-based abstractions. While effective, the resulting conservativeness grows with the time horizon, and scalability is limited by the need for domain splitting and refinement to balance precision and computational tractability.

Multi-step RL verification also admits multiple definitions, which differ in their treatment of dynamics, temporal specifications, and uncertainty, and which we review next.

\paragraph{Labeled MDPs and Temporal Logic Verification.} 
Many long-horizon RL objectives are naturally expressed using temporal logics such as linear temporal logic (LTL) \citep{LTL}, which enable explicit specification of properties over trajectories \citep{carr2021verifiable}. To reason about these, the environment is modeled as a labeled MDP, where states are annotated with atomic propositions via a labeling function $L$. The LTL specification $\varphi$ is translated into an automaton $M$, and verification is performed over the product of the policy-induced MDP and the automaton. Combined with reachability analysis and linearization-based over-approximations, this yields a conservative characterization of trajectories that may violate $\varphi$. This formulation substantially increases verification complexity, as the product MDP-automaton can be exponentially larger than the original system, making formal verification intractable for high-dimensional or long-horizon settings. Formally, given an initial state set $\mathcal{X}_0$ and an LTL formula $\varphi$, we define this type of verification in Def. \ref{def:ltl_rl_verification}.

\begin{tcolorbox}
\vspace{-0.1cm}
\begin{definition}[\textsc{LTL-Based Multi-Step RL Verification}]
\label{def:ltl_rl_verification}
\phantom{a}

{\bf Input}: A tuple $\mathcal{T} = \langle f, M, \mathcal{X}_0, \varphi \rangle$

{\bf Output}: $\texttt{violate} \iff \exists \tau \in \langle f, M, \mathcal{X}_0\rangle \;\vert\; \tau \not\models \varphi$
\end{definition}
\end{tcolorbox}

\paragraph{Neuro-Symbolic Certificates.}
On a different direction, neuro-symbolic approaches combine RL policies with symbolic safety certificates from control theory, most notably control barrier functions and Lyapunov functions \citep{CBF}. Shielding-based neuro-symbolic approaches can also guarantee safety at runtime, but have so far been demonstrated primarily in small-scale, discrete RL domains \citep{shielding}. In these paradigms, the policy proposes actions, while the certificate enforces safety by modifying or rejecting unsafe actions at runtime, enabling long-horizon guarantees.

When available analytically, such certificates provide strong model-based guarantees over continuous-time trajectories. In many RL settings, however, the safe set or certificate must be learned from data, leading to neural barrier or Lyapunov functions whose verification becomes a problem in its own right. Recent work focuses on certifying that learned certificates satisfy invariance or stability conditions over the relevant state space \citep{CBFverification,yang2024lyapunov}. While effective, neuro-symbolic methods shift rather than eliminate the verification burden, inheriting the scalability challenges of neural network verification and introducing additional modeling assumptions. Probabilistic variants further relax worst-case guarantees in favor of high-confidence safety certificates over long horizons \citep{ral}.

\paragraph{Probabilistic Multi-Step Verification.}
Even in the multi-step case, probabilistic verification relaxes worst-case guarantees by estimating the likelihood that an RL policy satisfies a temporal property over closed-loop trajectories. These methods achieve scalability by sampling trajectories or propagating probabilistic bounds through labeled MDPs, providing formal confidence guarantees rather than absolute safety. A representative example is the probabilistic reachability framework of \cite{Abate}, which computes bounds on the probability of satisfying temporal specifications in discrete-time stochastic systems.

\textit{Position within the taxonomy.} Multi-step verification methods extend step-wise verification along the temporal axis. They expose a fundamental trade-off between expressivity and scalability: richer behavioral guarantees require reasoning about policy-environment interaction over time, but quickly lead to state-space explosion. Understanding and managing this trade-off is central to RL verification and motivates the emerging problem we discuss next.

\subsection{Emerging Verification Formulations}
\label{sec:enumeration}
An emerging \emph{enumeration-based verification} paradigm (or neural network preimage analysis) (Def. \ref{def:enum_verification}) \emph{asks where in state space does the policy systematically fail?} In RL, where failures often arise from structured regions rather than isolated states, such region-level information supports policy debugging, targeted data collection, retraining, and deployment-time mitigation strategies \citep{matoba2020exact,CountingProve,INVPROP}.

\textit{Problem Formulation.}  Given a policy network and an unsafe output predicate, enumeration-based verification computes the set of inputs that lead to unsafe actions. This can be formally defined as follows:
\begin{tcolorbox}
\vspace{-0.1cm}
\begin{definition}[\textsc{AllDNN-Verification for RL}]
\label{def:enum_verification}
\phantom{a}

{\bf Input}: A tuple $\mathcal{T} = \langle f, \mathcal{X}, \mathcal{Y}\rangle$ 

{\bf Output}: $\Gamma(\mathcal{T}) = \{x \in \mathcal{X} \mid f(x) \in \mathcal{Y}\}$
\end{definition}
\end{tcolorbox}
Exact enumeration methods seek to compute $\Gamma(\mathcal{T})$ (i.e., the set of states that violate a property) precisely, for example, as a union of polytopes whose volume can be estimated. Recalling our navigation example, $\Gamma(\mathcal{T})$ may be represented by the red region of states highlighted in Fig.~\ref{fig:safety_prop} (b).
\begin{figure}[b!]
    \centering
    \includegraphics[width=.9\linewidth]{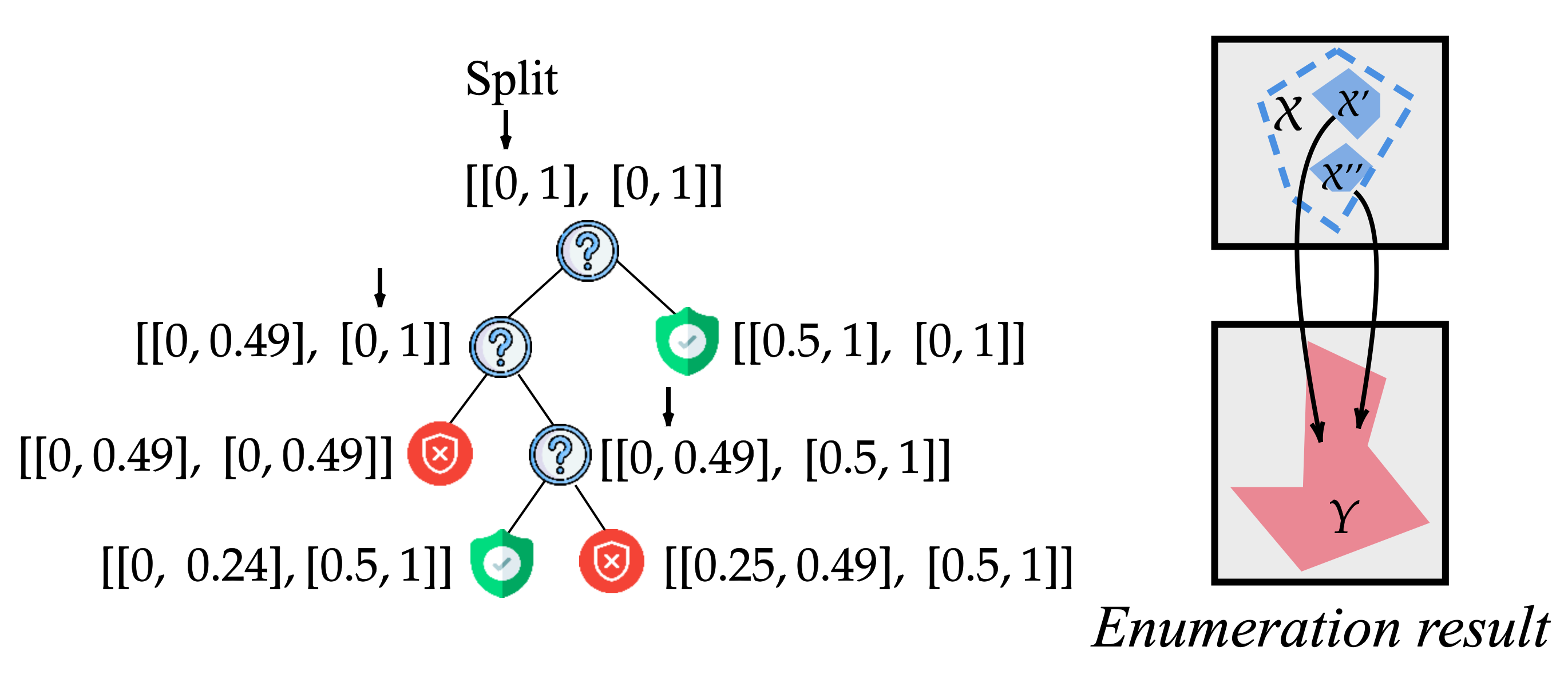}
    \vspace{-5pt}
    \caption{DNN preimage bound computation via BaB strategy. (Adapted from \citep{RFprove}.)}
    \vspace{-5pt}
    \label{fig:enum}
\end{figure}

In detail, enumeration-based approaches typically rely on branch-and-bound procedures that recursively partition the input space or activation patterns \citep{bab} as illustrated in Fig. \ref{fig:enum}. While enabling fine-grained safety analysis  \citep{matoba2020exact}, their worst-case complexity grows exponentially with network depth and input dimension, rendering exact enumeration \#P-hard \citep{CountingProve} and motivating the need for novel perspectives. 

\paragraph{Enumeration Approaches.} 
A first line of work proposes approximate preimage-bounding techniques based on linearization, which trade exactness for efficiency. Over-approximate methods, such as the backward analysis in \cite{INVPROP}, compute conservative outer bounds that contain all unsafe inputs, but still result in limited scalability for large RL policies.
To enable scalable solutions, recent frameworks either combine over- and under-approximation analyses, such as PREMAP \citep{zhang2025premap}, or adopt a probabilistic perspective, as in CountinProVe, $\epsilon$-ProVe, and RF-ProVe \citep{CountingProve,eProve,RFprove}. These approaches improve the scalability of preimage analysis for large RL policies, enabling practical region-level behavioral analysis.

\textit{Position within the taxonomy.}
Knowing \emph{where} unsafe decisions happen significantly pushes verification in the \emph{RL readiness} direction of our taxonomy. For example, preimage information can be used to: (i) guide additional training or data collection in high-risk regions, (ii) synthesize recovery or shielding mechanisms that avoid unsafe states, and (iii) assess the severity of residual risk by estimating the size or probability mass of unsafe regions. These use cases go beyond what binary verification can provide and align enumeration-based methods with practical RL workflows.

% \paragraph{Limitations and outlook.}
% While enumeration-based methods offer richer behavioral insight than binary verification, they remain computationally demanding and typically operate under step-wise abstractions. Extending enumeration to trajectory-level properties or integrating it tightly with environment dynamics remains an open challenge. Nevertheless, enumeration provides a powerful lens for understanding and improving RL policies, complementing both formal and probabilistic verification approaches discussed earlier.
\begin{figure}[b!]
    \centering
    \includegraphics[width=.75\linewidth]{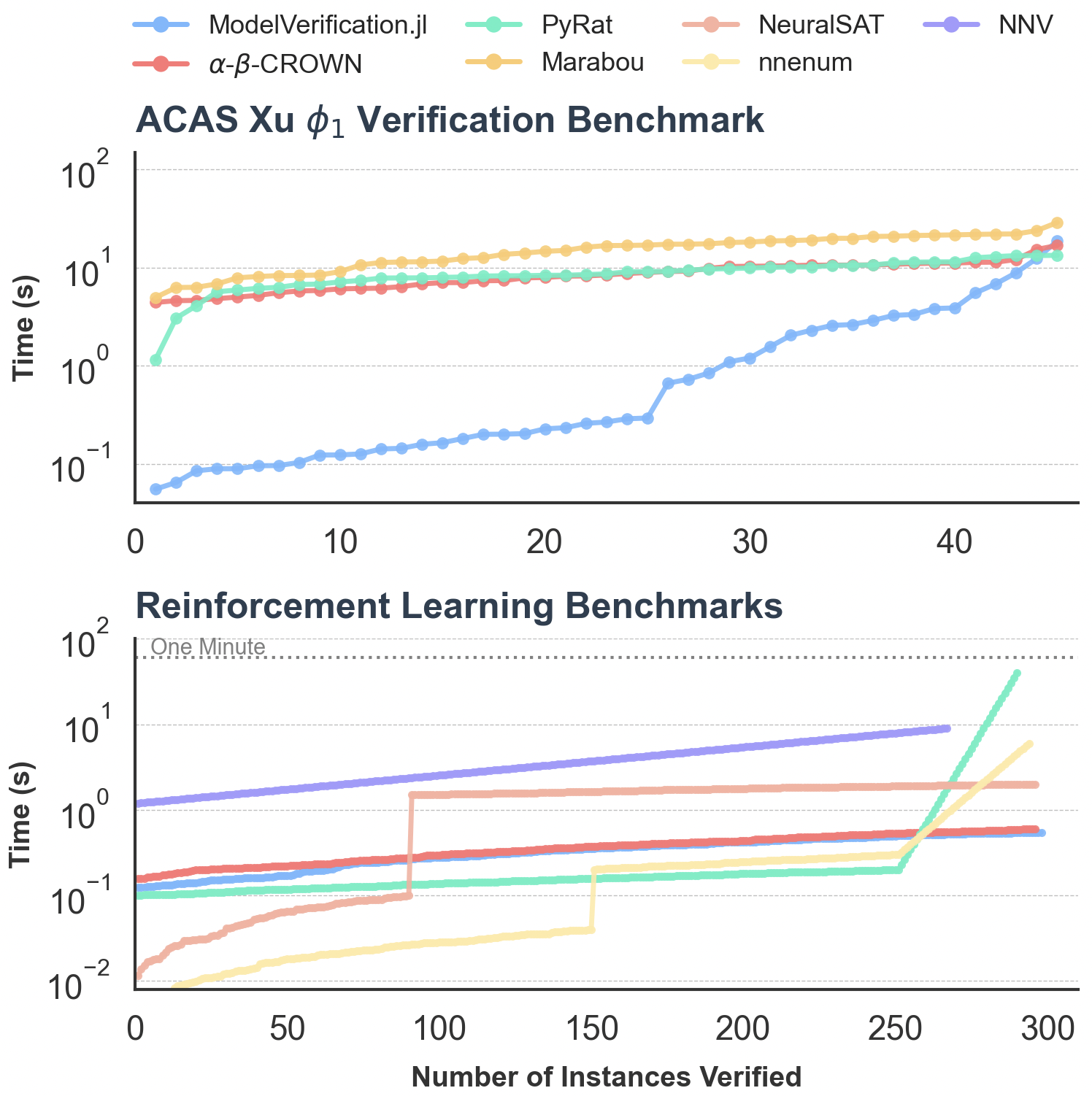}
    \vspace{-5pt}
    \caption{Runtime of RL verification tools. (Top) Performance on the ACAS Xu $\phi_1$ benchmark. (Bottom) Results on RL policy verification tasks. The x-axis denotes the number of verified properties (instances) and the y-axis the cumulative computation time.}
    \label{fig:benchmarks}
    \vspace{-5pt}
\end{figure}

\section{Verification Tool Benchmarking \& Selection}\label{sec:bench}

Verification of RL policies spans objectives ranging from local robustness under perception uncertainty to safety guarantees over closed-loop trajectories. Because these objectives differ in temporal scope, guarantee strength, and scalability, no single verification technique is universally applicable.

Fig. \ref{fig:benchmarks} and Table \ref{tab:tab_eval} complement our taxonomy with an empirical comparison of methods from (i) the international verification of neural networks competition (VNN-COMP) \citep{VNN-comp2023} and (ii) preimage approximation approaches from \cite{RFprove}. The benchmarks focus on open-loop sensor-noise robustness for discrete-action RL policies implemented as fully connected ReLU networks, trained on ACAS \citep{ACAS}, Dubin’s rejoin, CartPole, and LunarLander from OpenAI Gym \citep{gym} tasks. 

\paragraph{Peformance Analysis.}
Many methods perform similarly on ACAS Xu, but their scalability diverges on RL benchmarks. For step-wise verification, bound-propagation methods consistently outperform exact solvers as the number of instances grows, while relaxation-based approaches better handle the dimensionality and repetition of RL policy analysis. $\alpha,\beta$-CROWN and ModelVerification.jl exhibit promising scaling across benchmarks, highlighting the effectiveness of tight linear relaxations and efficient branching heuristics.

For multi-step verification, the lack of standardized benchmarks remains an open challenge, reflecting fundamental difficulties in modeling dynamics, uncertainty, and temporal specifications rather than an omission of this survey. Nevertheless, scalability is the dominant bottleneck: reasoning over closed-loop trajectories rapidly expands the state space, limiting exact or worst-case guarantees. Control-theoretic methods based on Lyapunov or control barrier functions provide strong guarantees when applicable, but rely on additional modeling assumptions that constrain generality.
\begin{table}[t]
    \tiny
    \centering
    \begin{tabular}{lcccc}
    \textbf{Method} & \textbf{Task} & \textbf{Coverage} & \textbf{Error (\%)} & \textbf{Time}\\
    \toprule
    Exact & VCAS & 100\% & 0\% & 6352.21s\\
    PREMAP & VCAS & \textbf{90.8\%} & 0\% & 12.8s\\
    $\varepsilon$-ProVe & VCAS & 90.48\% & 0.02\% & 0.65s\\
    RF-ProVe & VCAS & 90.5\% & 0.06\% & 0.3s\\
    \midrule
    PREMAP & Cartpole & 75.5\% & 0\% & 32.37s\\
    $\varepsilon$-ProVe & Cartpole & 76.47\% & 0.27\% & 2s\\
    RF-ProVe & Cartpole & \textbf{76.8\%} & 0.3\% & 4.5s\\
    \midrule
    PREMAP & Lunarlander & 75.1\% & 0\% & 85.42s\\
    $\varepsilon$-ProVe & Lunarlander & 76.51\% & 0.5\% & 12.2s\\
    RF-ProVe & Lunarlander & \textbf{75.63\%} & 0.3\% & 59s\\
    \midrule
    PREMAP & Dubin's rejoin & 78.7\% & 0\% & 656.47s\\
    $\varepsilon$-ProVe & Dubin's rejoin & 85.02\% & 0.3\% & 260.23s\\
    RF-ProVe & Dubin's rejoin & \textbf{90.08\%} & 0.3\% & 66s\\
    \bottomrule
    \end{tabular}
    \caption{Enumeration-based (preimage) evaluation from \cite{RFprove}. We report coverage of the unsafe preimage, approximation error, and runtime, highlighting the trade-offs between exactness, scalability, and computational efficiency.}
    \label{tab:tab_eval}
    \vspace{-3mm}
\end{table}

%Finally, enumeration-based methods deliver the strongest guarantees by explicitly characterizing unsafe regions, yet both the taxonomy and experimental results confirm that their computational cost grows sharply with network depth and input dimension. Overall, the results position enumeration as a powerful but specialized tool, best suited for detailed analysis in low-dimensional settings rather than as a scalable solution for general RL policy verification.
Finally, enumeration-based methods provide the strongest guarantees by explicitly characterizing unsafe regions, but their cost grows rapidly with network depth and input dimension. Exact approaches such as \citep{matoba2020exact} do not scale to larger networks or complex RL specifications, often producing highly fragmented polytope representations that exhaust memory and hinder interpretability. Approximate methods \citep{zhang2025premap}, particularly probabilistic approaches such as $\epsilon$-ProVe and RF-ProVe, substantially improve scalability by relaxing soundness in exchange for compact, informative representations better suited to downstream tasks such as safe recovery and explanation.

These results suggest that scalability, architectural coverage, and tooling maturity are tightly coupled dimensions rather than independent factors.

\paragraph{Verification Tool Selection.} Tool selection depends on whether a method’s assumptions and guarantees match the RL setting. In the following, we summarize practical guidelines for tool selection in RL: (i) \emph{Local robustness or action-level safety}: Over-approximate reachability-based methods are often effective when certifying safety over neighborhoods of states, especially under epistemic uncertainty. (ii) \emph{Falsification and debugging}: SAT-based solvers are well suited when concrete counterexamples are desired to diagnose policy failure modes. (iii) \emph{High-dimensional spaces or large-scale evaluation}: Relaxation-based and probabilistic methods tend to scale better when verification must be performed repeatedly across many states \citep{VNN-comp2023}. (iv) \emph{Region-level risk analysis}: Enumeration-based tools are most appropriate when the goal is to localize and quantify unsafe regions of the state space rather than obtain a binary verdict. (v) \emph{Long-horizon or deployment-oriented safety}: Multi-step or neuro-symbolic approaches are preferable when guarantees must hold over closed-loop trajectories.

Taken together, the taxonomy and empirical results provide practical guidance for selecting verification methods that are well matched to the structure, scale, and verification objectives of a given RL application, while making explicit the assumptions under which each method is most effective.

\section{Open Challenges}\label{sec:open_challenges}
Several challenges remain in post-training RL verification. We focus on two settings that most clearly expose current limitations: \emph{history-dependent policies}, typically implemented via recurrent neural networks (RNNs), and \emph{multi-agent RL} (MARL). In recurrent RL, verification must reason over feasible hidden-state trajectories induced by interaction histories, while in MARL this challenge extends to agents whose local observations and histories may jointly induce system-level violations. We also discuss orthogonal challenges spanning modern policy architectures and temporal scopes. Together, these issues highlight a fundamental mismatch between classical neural network verification and modern RL systems, where correctness depends on policy-induced histories, interactions, and closed-loop dynamics rather than static inputs.

\paragraph{Verifying Recurrent Policies.}
A key challenge in RL verification arises from recurrent policies in partially observable environments. Unlike feedforward models, RNNs induce history-dependent behavior via latent hidden states, making safety and robustness depend on the set of feasible hidden states. Most existing methods are step-wise and memoryless, reducing RNN verification to bounded input perturbations over a fixed unrolling horizon; an abstraction that captures local robustness but fails to reflect uncertainty arising from the agent’s internal memory under partial observability.

\textit{Problem Formulation.} To incorporate recurrent policies into our unified taxonomy, we formalize the problem of as reasoning over uncertainty in the policy’s internal state (Def \ref{def:drl_rnn_safety}). Given an RNN-based policy $f$, a set of histories $\mathcal{H}$, a current observation $o$, and a postcondition $\mathcal{Y}$ encoding unsafe behavior as in step-wise SAT verification (Def. \ref{def:sat_rl_verification}), the verification problem can be formulated to ask whether there exists a feasible hidden state $h \in \mathcal{H}^*$, corresponding to a realizable interaction history, such that $f(h, o) \in \mathcal{Y}$.
This formulation inherently spans multi-step temporal reasoning.

\begin{tcolorbox}
\vspace{-0.1cm}
\begin{definition}[\textsc{RNN-Verification} for RL]\label{def:drl_rnn_safety}
\phantom{a}

    {\bf Input}: A tuple $\mathcal{T}=\langle f, \mathcal{H}, o, \mathcal{Y} \rangle$
    
    {\bf Output}: $\texttt{violate} \iff \exists h \in \mathcal{H}^* \;\vert\; f(h, o) \in \mathcal{Y}$
\end{definition}
\end{tcolorbox}

Prior work on verifying recurrent policies either relies on discretizing RNNs into finite-state abstractions~\citep{carr2021verifiable} or on reachability analysis assuming known environment dynamics~\citep{everett2021reachability}. Despite their conceptual appeal, these approaches impose strong modeling assumptions and may suffer from poor scalability, as the abstract state space grows exponentially with the hidden-state dimension or the planning horizon.
Crucially, a key challenge is that the feasible hidden-state set $\mathcal{H}^*$ is highly structured and policy-dependent. Standard reachability or interval-based abstractions typically over-approximate this set and may admit unrealizable hidden states in their representations, as they lack access to a \textit{feasibility oracle} capable of distinguishing hidden states induced by realizable interaction histories from infeasible ones. As a result, the abstract feasibility space may contain spurious counterexamples or yield overly conservative verification guarantees. On top of this aspect, even assuming access to such a feasibility oracle, exactly characterizing unsafe hidden states is computationally intractable, subsuming \#P-hard counting and preimage problems even in the feedforward case~\citep{CountingProve}.

This exposes a mismatch, where behavior depends on policy-induced histories rather than single inputs. Open questions include: \emph{how to define verification objectives faithful to history-dependent decision making in RL? How to characterize feasible histories without explicit enumeration or overly conservative over-approximation?} Progress will require frameworks that reason over feasible histories, incorporate probabilistic guarantees when worst-case analysis is intractable, and better align formal methods with RL-specific uncertainty and execution semantics.

\paragraph{Verifying Multi-Agent Systems.}
Safety and correctness in MARL emerge from interactions among multiple agents. Verification thus depends on the learning and execution paradigm (e.g., cooperative or adversarial settings, centralized or decentralized policies), and varying coordination assumptions. These challenges are compounded by the exponential growth of the joint state-action space with the number of agents, and the need for recurrent architectures to enable coordination. Hence, we focus on the common fully cooperative setting with decentralized execution, typically realized via centralized training with decentralized execution \citep{papoudakis2021benchmarking}. A team of $\mathcal{N}$ agents must coordinate to achieve a shared objective, where each agent $i$ executes an RNN-based policy $f_i$ based solely on its local observation $o_i$ and hidden state $h_i \in \mathcal{H}_i$ at deployment.

\textit{Problem Formulation.}
Considering our taxonomy, we formalize cooperative MARL verification (Def \ref{def:marl_rnn_safety}). Given agent-specific verification problems $\mathbf{T} = \{\mathcal{T}_1, \ldots, \mathcal{T}_\mathcal{N}\}$, with $\mathcal{T}_i = \langle f_i, \mathcal{H}_i, o_i, \mathcal{Y} \rangle$, a team-level violation occurs if at least one agent admits a feasible hidden state $h_i \in \mathcal{H}_i^*$ such that its local policy violates the cooperative behavioral property (encoded in $\mathcal{Y}$). This formulation operates over a multi-step horizon induced by decentralized interaction.

\begin{tcolorbox}
\vspace{-0.1cm}
\begin{definition}[\textsc{RNN-Verification} for cooperative MARL]\label{def:marl_rnn_safety}
\phantom{a}

    {\bf Input}: $\mathbf{T} = \{ \mathcal{T}_1, \ldots, \mathcal{T}_\mathcal{N} \}$, with $\mathcal{T}_i = \langle f_i, \mathcal{H}_i, o_i, \mathcal{Y} \rangle$
    
    {\bf Output}: $\texttt{violate} \iff \exists i \in \{1, \ldots, \mathcal{N}\},\\  \text{\hspace{3.5cm}} \exists h_i \in \mathcal{H}_i^* \;\vert\; f_i(h_i, o_i) \in \mathcal{Y}$
\end{definition}
\end{tcolorbox}
Extending this formulation to adversarial or mixed cooperative-competitive MARL introduces additional challenges, as verification must account for strategic opponents and worst-case behaviors, fundamentally altering semantics and tractability. Under decentralized execution, agent-wise verification can be performed independently, with system-level safety determined by the worst-performing agent, enabling parallel analysis without added conservatism. However, this decomposition raises open questions, such as: \emph{when is agent-wise verification sufficient to guarantee team-level safety? How should verification account for implicit coordination encoded in agents’ recurrent hidden states?}

More broadly, MARL exposes a gap between classical verification and decentralized learning, motivating frameworks that reason about interaction-induced uncertainty, emergent behavior, and history-dependent coordination while balancing soundness and scalability.

\paragraph{Orthogonal Challenges.}
Beyond these settings, several orthogonal challenges further complicate RL verification across paradigms, objectives, and temporal horizons. (i) \emph{Environment uncertainty and closed-loop verification}. Most methods assume known or accurately approximated dynamics, whereas real-world RL systems operate under model mismatch, disturbances, and non-stationarity, raising questions about whether guarantees should apply to the policy, the closed-loop system, or uncertainty sets over environments, and how such guarantees can be provided tractably. (ii) \emph{Long-horizon reasoning and rare events}. Many safety-critical failures arise from low-probability events over long horizons, which worst-case methods either miss or over-approximate excessively, making it difficult to capture risk accumulation, infinite-horizon behavior, and rare-event probabilities at scale. (iii) \emph{Modern architectures} such as Transformers, the verification challenge is twofold: architectural, because attention and softmax violate piecewise-linear assumptions, and semantic, because these models reason over variable-length histories via attention rather than explicit recurrence. As a result, existing abstractions struggle to scale and to faithfully capture temporal context, leaving this gap an open direction.

These challenges highlight that RL verification is not merely a scaling problem but a conceptual one: correctness must be defined in terms of interactions, temporal dependence, and decentralized decisions.

\section{Conclusion}\label{sec:conclusion}

This survey presented a unified view of post-training RL verification, organizing a fragmented literature along three axes: verification paradigm, objective, and temporal scope. Our taxonomy clarifies the assumptions, guarantees, and scalability trade-offs of existing methods, highlighting the tension between strong guarantees and computational tractability.

More broadly, our analysis shows that RL verification requires a conceptual shift beyond classical neural network verification, as correctness depends on closed-loop interaction, temporal dependence, and increasingly on history-dependent and decentralized decision making. Addressing these challenges (particularly for recurrent policies, multi-agent systems, long horizons, and modern architectures) will require verification frameworks that better align with RL execution semantics while balancing soundness and scalability.

\clearpage

% \section*{Acknowledgements}
% This work was supported in part by the ``Fondo Italiano per la Scienza" project (FIS-2024-05614), and in part by the Austrian Science Fund (FWF) project 10.55776/ESP1944725, the Vienna Science and Technology Fund (WWTF) under Grant ICT22-023 (TAIGER) and the European Union, RobustifAI project, ID 101212818. The authors thank the reviewers for their insightful and constructive feedback, which has substantially improved the quality of this work.

%% The file named.bst is a bibliography style file for BibTeX 0.99c
\bibliographystyle{named}
\bibliography{ijcai26}

\end{document}